\RequirePackage[svgnames,table]{xcolor}
\documentclass[11pt,letterpaper,logo]{railstyle}
\usepackage[numbers,sort&compress,square]{natbib}
\providecommand{\papertitle}{Recovering What Matters, Transferring Only What Is Reliable: Toward Precise Reference-Based Super-Resolution}
\usepackage{graphicx}
\usepackage{amsmath}
\usepackage{amssymb}
\usepackage{booktabs}
\usepackage{xcolor}
\usepackage{bbm}
\newcommand{\best}[1]{\textbf{\textcolor{red}{#1}}}
\newcommand{\second}[1]{\textbf{\textcolor{blue}{#1}}}
\usepackage{multirow}
\usepackage{colortbl}
\usepackage{adjustbox}
\usepackage{stfloats}

\usepackage{xspace}
\usepackage{caption}
\usepackage{subcaption}
\usepackage{etoolbox}
\definecolor{cvprblue}{rgb}{0.21,0.49,0.74}
\makeatletter
\DeclareRobustCommand\onedot{\futurelet\@let@token\@onedot}
\def\@onedot{\ifx\@let@token.\else.\null\fi\xspace}

\AtEndPreamble{%
  \@ifpackageloaded{hyperref}{}{\usepackage[breaklinks,colorlinks,allcolors=cvprblue]{hyperref}}%
  \@ifpackageloaded{cleveref}{}{\usepackage[capitalize]{cleveref}}%
  \crefname{section}{Sec.}{Secs.}\Crefname{section}{Section}{Sections}%
  \Crefname{table}{Table}{Tables}\crefname{table}{Tab.}{Tabs.}%
}
\makeatother

\usepackage{amsmath,amsfonts,bm}

\def\eqref#1{equation~\ref{#1}}

\def\1{\bm{1}}

\def\mEps{{\bm{\epsilon}}}
\def\vzero{{\bm{0}}}
\def\vone{{\bm{1}}}
\def\vmu{{\bm{\mu}}}

\def\vk{{\bm{k}}}

\def\vq{{\bm{q}}}

\def\mC{{\bm{C}}}

\def\mH{{\bm{H}}}
\def\mI{{\bm{I}}}

\def\mX{{\bm{X}}}

\def\mZ{{\bm{Z}}}

\DeclareMathAlphabet{\mathsfit}{\encodingdefault}{\sfdefault}{m}{sl}
\SetMathAlphabet{\mathsfit}{bold}{\encodingdefault}{\sfdefault}{bx}{n}

\newcommand{\softmax}{\mathrm{softmax}}

\newcommand{\teasercaption}{Our method reads fine details from matching references and resists mismatched ones. \emph{Top left:} restoration quality on CUFED5 and WR-SR and content recovery on VeriRef; farther from the center is better. \emph{Top right:} ArcFace similarity to the ground-truth identity under correct and wrong references. \emph{Bottom:} compared with Ada-RefSR~\cite{wang2026adarefsr} (ICLR~2026), our method recovers legible text, decodable codes, and identity cues from the reference, and finer details in general scenes.}

\newcommand{\abstracttext}{Reference-based super-resolution (RefSR) aims to recover missing details from high-quality references. However, existing methods face two fundamental challenges. {First, perceptual plausibility does not guarantee content fidelity}. Existing methods often produce visually convincing details without faithfully recovering reference-specific information, such as text, codes, and identity cues. {Second, faithful reference reading does not guarantee reliable information transfer}. Increasing reference utilization also exposes the restoration process to misleading content, as global semantic relevance does not necessarily imply local correspondence consistency. To address these challenges, we propose \textbf{PreciseRef}, a reference-aware generative framework for faithful and reliable RefSR. At its core, Asymmetric Twin-Attention (ATA) allows target and reference images to participate in a unified attention space while the reference keeps its own shape. By letting their image tokens interact directly through joint attention, ATA facilitates fine-grained reference information retrieval without explicit image alignment. To ensure reliable information transfer, we introduce Dual-Scale Reference Gating (DRG), which combines Semantic Relevance Gating (SRG) to filter globally irrelevant reference context and Bidirectional Consistency Gating (BCG) to suppress unreliable local correspondences. Furthermore, we establish VeriRef, a benchmark evaluating exact reference-dependent content recovery and robustness to mismatched references. Extensive experiments show that our method achieves 85.6\% character accuracy and 76.3\% exact code decoding, compared with 17.4\% and 0\% for existing methods, while effectively suppressing erroneous reference transfer. It also achieves state-of-the-art perceptual quality on CUFED5 and WR-SR.}

\newcommand{\duallogo}[1]{\raisebox{-0.5\height}{\includegraphics[height=#1]{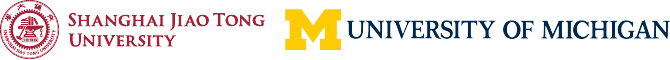}}}

\newcommand{\teaserblock}{%
  \par\vspace{-3mm}\noindent
  \begin{minipage}{\textwidth}\centering%
    \includegraphics[width=0.98\textwidth]{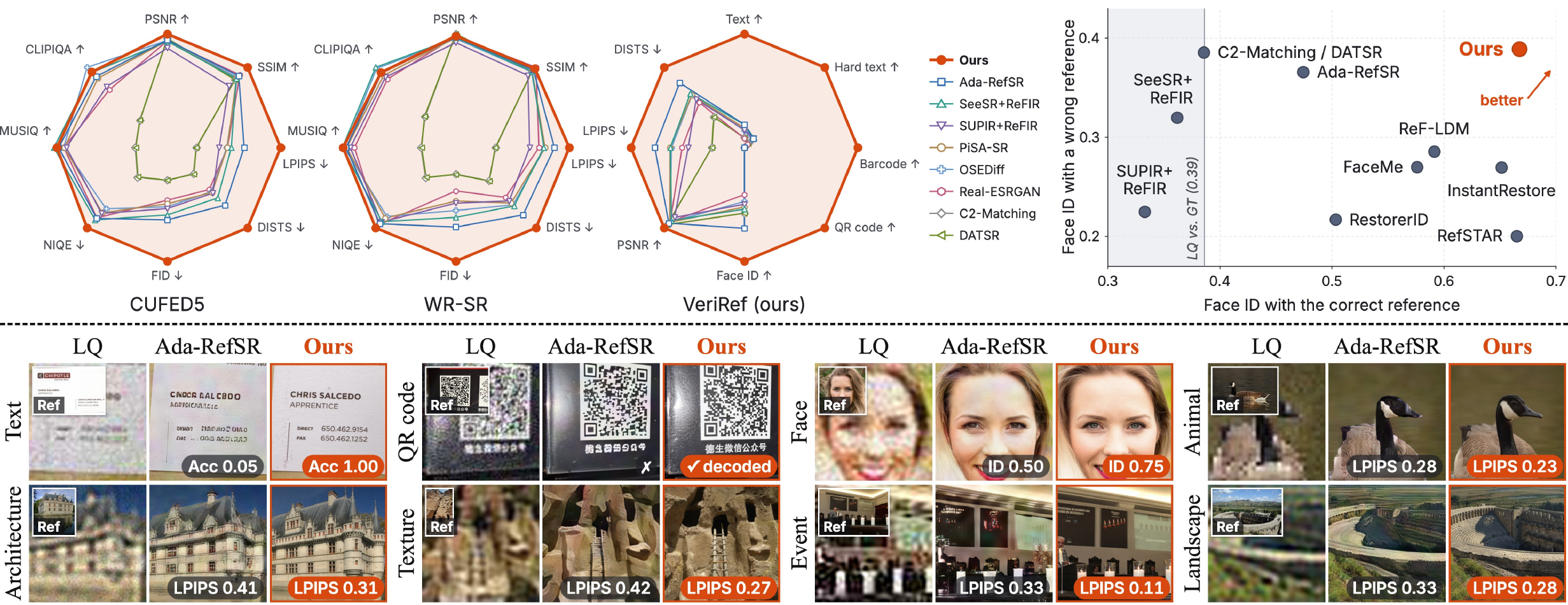}
    \captionsetup{type=figure,hypcap=false,skip=3pt,font=footnotesize}
    \captionof{figure}{\teasercaption}
    \label{fig:teaser}
  \end{minipage}\par\vspace{\baselineskip}}

\newlength{\sfwidth}
\AtBeginDocument{\setlength{\sfwidth}{\textwidth}}
\makeatletter
\let\rsr@figure\figure \let\rsr@endfigure\endfigure
\renewenvironment{figure}[1][tbp]{\rsr@figure[#1]\centering\begin{minipage}{\sfwidth}}{\end{minipage}\rsr@endfigure}
\let\rsr@table\table \let\rsr@endtable\endtable
\renewenvironment{table}[1][tbp]{\rsr@table[#1]\centering\begin{minipage}{\sfwidth}}{\end{minipage}\rsr@endtable}
\makeatother

\title{\papertitle}
\runningtitle{Toward Precise Reference-Based Super-Resolution}
\author[1]{Sihan Ji}
\author[2]{Wenxin Chen}
\author[3]{Yue Shi}
\author[1,\dag]{Jiezhang Cao}
\correspondingauthor{Jiezhang Cao (\texttt{caojiezhang@sjtu.edu.cn})}
\affil[1]{Shanghai Jiao Tong University}
\affil[2]{University of Michigan}
\affil[3]{Shanghai Artificial Intelligence Laboratory}
\begin{abstract}
\abstracttext
\end{abstract}
\begin{document}
\maketitle
\teaserblock
\section{Introduction}
\label{sec:intro}

Reference-based super-resolution (RefSR) aims to recover details lost in a low-quality (LQ) input by leveraging a high-quality reference image. Its unique advantage lies in retrieving information no longer discernible from the degraded input alone. For example, when degradation erases text, product codes, or facial identity cues, generative priors may synthesize visually plausible details but cannot reliably recover the original content. Such information must be retrieved from a reference depicting the same content. However, existing RefSR methods face two challenges.

\begin{figure}[t]
    \centering
    \includegraphics[width=\linewidth]{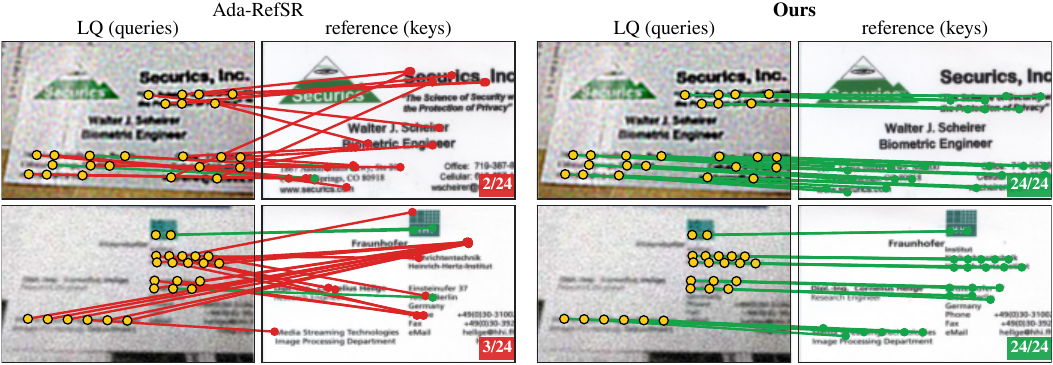}
    \caption{Lines link LQ queries to their peak attention on the reference (green: correct, red: wrong). Ours: 48/48; Ada-RefSR: 5/48.}
    \label{fig:attn_match}
\end{figure}

First, perceptual plausibility does not guarantee content fidelity, as visually convincing details may not faithfully recover reference-specific information. Existing methods, from feature-matching networks~\cite{zhang2019srntt,jiang2021c2matching,cao2022datsr} to diffusion restorers with reference branches~\cite{guo2024refir,wang2026adarefsr}, produce convincing details but rarely recover exact reference-specific content. To understand this limitation, we conduct controlled comparisons using the same FLUX~\cite{fluxfill} backbone, training data, and schedule. A frozen reference extractor achieves only 6.3\% code decoding, while separate-softmax attention and one-way injection reach 60--65\%. In contrast, joint attention over reference and target tokens improves decoding to 71.3\%, revealing the reference interaction pathway as a critical bottleneck for fine-grained information retrieval. From \cref{fig:attn_match}, our joint attention locates the matching reference content for almost all LQ queries; Ada-RefSR rarely does.

Second, faithful reference reading does not guarantee reliable information transfer, as references may also introduce misleading content. This problem manifests at two levels: an unrelated reference may introduce incorrect global semantics, such as imposing the textures and appearance of a different scene on the whole image, while visually similar but mismatched local regions may cause erroneous detail transfer, such as copying digits from another code. More fundamentally, global semantic relevance does not guarantee local correspondence consistency: a semantically relevant reference may still contain misleading local details. Our experiments demonstrate the importance of addressing both aspects. Filtering local correspondences by bidirectional consistency reduces incorrect digit copying from 21.8\% to 10.9\%, while additionally gating global context by semantic relevance removes 66\% and 64\% of the excess LPIPS errors induced by mismatched references on CUFED5 and WR-SR, respectively. These findings motivate selective reference utilization at both levels.

\begin{figure*}[t]
    \centering
    \includegraphics[width=\textwidth]{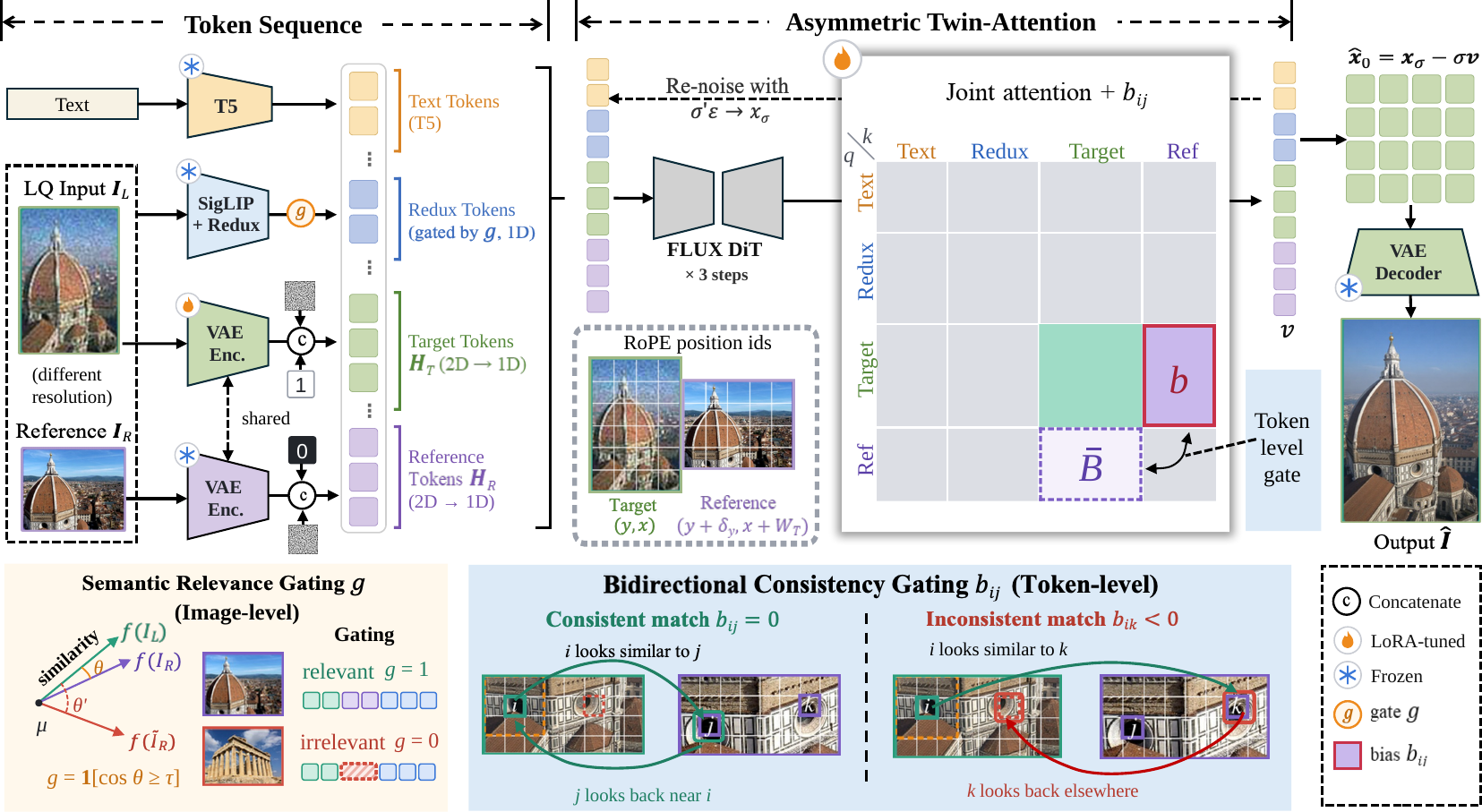}
   \caption{Overview of PreciseRef. \emph{Top:} Asymmetric Twin-Attention (ATA) places the target and reference tokens $\mH_T$ and $\mH_R$ (\cref{eq:tokens}) in one joint-attention sequence; the reference keeps its own shape and sits at $(y+\delta_y,\,x+W_T)$ in RoPE coordinates, where $W_T$ is the target width. \emph{Bottom:} Dual-Scale Reference Gating (DRG). Semantic Relevance Gating (SRG, \cref{eq:gate}) drops the Redux tokens of an irrelevant reference, and Bidirectional Consistency Gating (BCG, \cref{eq:mcb}) derives $b_{ij}$ from the reverse attention $\bar{B}$ to suppress inconsistent matches.}
   \label{fig:pipeline}
\end{figure*}

To address these challenges, we propose PreciseRef, a reference-aware generative framework combining Asymmetric Twin-Attention (ATA) and Dual-Scale Reference Gating (DRG). ATA jointly processes reference and target tokens within a unified attention space while preserving the reference’s own aspect ratio and token count, enabling fine-grained reference retrieval without explicit image alignment. To ensure reliable information transfer, DRG regulates reference information at two complementary levels: Semantic Relevance Gating (SRG) uses centered SigLIP~\cite{siglip} features to filter globally irrelevant reference context, while Bidirectional Consistency Gating (BCG) exploits reverse attention consistency to suppress unreliable local correspondences. Both gates require no additional learned parameters or mismatched-reference training data. Furthermore, we establish VeriRef, a benchmark featuring paired correct and mismatched references across text, codes, and faces, to jointly evaluate precise content recovery and robustness to misleading information. Extensive experiments show that PreciseRef achieves 85.6\% character accuracy and 76.3\% exact code decoding, compared with 17.4\% and 0\%, respectively, for the best existing method. It also effectively suppresses erroneous reference transfer and achieves state-of-the-art perceptual quality on CUFED5~\cite{zhang2019srntt} and WR-SR~\cite{jiang2021c2matching}, with the best LPIPS, DISTS, and FID among the compared methods.

Our contributions are summarized as follows:
\begin{itemize}
\item Different from existing approaches that rely on separate reference feature extraction, explicit correspondence estimation, or feature warping, we introduce Asymmetric Twin-Attention to let reference and target tokens interact directly within a unified attention space while preserving the reference’s own aspect ratio and token count.
\vspace{1mm}
\item We develop Dual-Scale Reference Gating to balance effective reference utilization and reliable information transfer through image-level semantic relevance and token-level bidirectional consistency.   It admits relevant global context and suppresses unreliable local correspondences without additional learned parameters or mismatched-reference training.
\vspace{1mm}
\item Unlike conventional RefSR benchmarks that primarily assess perceptual quality, we reveal the overlooked gap between visually plausible restoration and exact reference-specific content recovery. We establish VeriRef, featuring paired correct and mismatched references across text, codes, and faces, to jointly evaluate genuine reference utilization and resistance to misleading information.
\end{itemize}

\section{Related Work}
\label{sec:related}

\paragraph{Reference-based Super-resolution (RefSR).}
RefSR restores a low-quality (LQ) image with a high-quality reference that shares its content.
Classical methods match features between the LQ input and the reference and transfer the matched textures~\cite{zheng2018crossnet,zhang2019srntt,yang2020ttsr,lu2021masasr}, as in C2-Matching~\cite{jiang2021c2matching} and DATSR~\cite{cao2022datsr}.
Recent methods condition a pretrained diffusion restorer on the reference, such as ReFIR~\cite{guo2024refir} and Ada-RefSR~\cite{wang2026adarefsr}, and face-specific methods condition on photographs of the same identity, such as ReF-LDM~\cite{refldm}, RestorerID~\cite{restorerid}, InstantRestore~\cite{instantrestore}, FaceMe~\cite{faceme}, and RefSTAR~\cite{refstar}.
Some of them share attention across images: InstantRestore~\cite{instantrestore} lets target queries attend to reference keys and values; others pass the reference through a matching module or an injection branch.
We instead process target and reference tokens jointly in every block of FLUX, so that its pretrained joint attention serves as the matching mechanism, and compare this with frozen-extractor, separate-softmax, and one-way variants under a matched setup (\cref{sec:inject}).

\vspace{-4mm}
\paragraph{Diffusion Priors for Single-image SR.}
Pretrained text-to-image diffusion models~\cite{ldm,sdxl} have become the dominant prior for real-world super-resolution, in multi-step restorers such as StableSR~\cite{stablesr}, DiffBIR~\cite{diffbir}, SUPIR~\cite{supir}, and SeeSR~\cite{seesr}, and one-step restorers such as OSEDiff~\cite{osediff}, S3Diff~\cite{s3diff}, and PiSA-SR~\cite{pisasr}.
They restore from the LQ input alone: erased content such as small text, codes, or identity cannot be inferred and needs a reference.

\vspace{-4mm}
\paragraph{In-context Generation and Joint-Attention in Latent Space.}
Diffusion transformers process text and image tokens with joint attention~\cite{mmdit}, so a condition image can be supplied as extra tokens in the same sequence, as in IC-LoRA~\cite{incontextlora}, OminiControl~\cite{ominicontrol}, Diptych Prompting~\cite{diptychprompting}, and FLUX.1 Kontext~\cite{kontext}.
These methods target synthesis and editing; we adapt the mechanism to restoration, where the output must stay faithful to a degraded input and the reference may not match it.
\section{Method}
\label{sec:method}

Given a degraded target $\mI_L$ and a reference $\mI_R$, reference-based super-resolution reconstructs a high-quality image $\hat{\mI}$ that is faithful to the target scene. The reference matters only for content that the LQ input no longer determines, and this places two requirements on a model. It must \emph{read} the reference: fine structures such as characters, code bars, and facial details have to be copied at the pixel level from wherever they appear in the reference. It must also \emph{reject} the reference where it does not match the target, since a model that copies faithfully also copies wrong content faithfully. ATA (\cref{sec:ata}) addresses the first requirement and DRG (\cref{sec:drg}) the second.

\vspace{-4mm}
\paragraph{Overview.} We build on FLUX.1-Fill~\cite{fluxfill}, a transformer whose 19 double-stream and 38 single-stream blocks all apply joint attention over text and image tokens. As shown in \cref{fig:pipeline}, the reference reaches the model through two pathways. In the \emph{spatial} pathway, ATA places the reference image tokens in the same sequence as the target tokens, so that every target token can read reference content directly. In the \emph{global} pathway, Redux~\cite{redux} tokens, SigLIP~\cite{siglip} image embeddings projected into the text stream, are appended for both the LQ input and the reference to convey what the reference depicts. DRG places one gate on each pathway.

\subsection{Asymmetric Twin-Attention}
\label{sec:ata}

\textbf{Motivation.} Existing RefSR methods pass the reference through a dedicated route, a matching module~\cite{jiang2021c2matching,cao2022datsr} or an injection branch~\cite{guo2024refir,wang2026adarefsr}, which recovers less fine content than joint attention in our controlled comparison (\cref{sec:inject}). We therefore let the reference enter the same attention as the target, so that the backbone itself decides where to read.

\vspace{-4mm}
\paragraph{Twin tokens.} With the frozen VAE encoder $E$, a separately adapted LQ encoder $E_L$, $\mZ_R=E(\mI_R)$, and $\mC_L=E_L(\mI_L)$, each image token carries a noisy latent, a condition, and a mask:
\begin{equation}
\begin{aligned}
\mH_T &= [\mX_\sigma,\,\mC_L,\,\vone],\\
\mH_R &= [(1-\sigma)\mZ_R+\sigma\mEps_R,\,\mZ_R,\,\vzero].
\end{aligned}
\label{eq:tokens}
\end{equation}
The target is generated from standard Gaussian noise $\mX_1$, while the clean reference stays visible through its condition channels at every noise level; only the target tokens are decoded and supervised. Because the two images share every attention operation, each target token can retrieve any reference token in every block, and the retrieved features enter the target in the same representation space the backbone uses for its own content. No correspondence network, warping, or fusion branch is involved; the pretrained attention of FLUX serves as the matching mechanism.

\vspace{-4mm}
\paragraph{Asymmetric shapes.} The reference keeps its own aspect ratio and token count rather than being cropped, padded, or resized to the target's shape. It is rescaled by the same factor as the target, preserving the relative scale of content, with its area capped at that of the target, so the joint sequence is at most twice the single-image length. For rotary position encoding~\cite{rope}, the target keeps its single-image coordinates and the reference is placed to its right, centered vertically.

\begin{figure}[t]
    \centering
    \includegraphics[width=\linewidth]{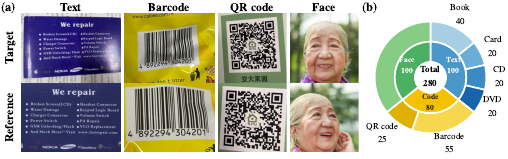}
    \caption{The VeriRef benchmark. (a) Example targets with correct references. (b) Composition of the 280 samples.}
    \label{fig:benchmark}
\end{figure}

\begin{figure*}[t]
\centering
\captionof{table}{Content recovery on VeriRef. \emph{Left:} text and codes with correct references; Decode is the rate of codes decoded to the exact payload. \emph{Right:} faces; ArcFace similarity to the ground truth with a correct (ID) or wrong (ID$_{\text{wrong}}$) reference, and to the distractor (ID leak$_{\text{wrong}}$). * denotes reference-based methods. \best{Red} and \second{blue}: best and second best; zero scores are not ranked.}
\label{tab:content}
\vspace{-1.5mm}
\newcommand{\zr}[1]{\textcolor{gray}{#1}}
\setlength{\tabcolsep}{5pt}
\adjustbox{max width=\textwidth}{%
\begin{tabular}{l|cccc|l|ccccc}
\toprule
Method & Text acc.$\uparrow$ & Decode$\uparrow$ & PSNR$\uparrow$ & LPIPS$\downarrow$ & Method & ID$\uparrow$ & ID$_{\text{wrong}}\uparrow$ & ID leak$_{\text{wrong}}\downarrow$ & PSNR$\uparrow$ & LPIPS$\downarrow$ \\
\midrule
C2-Matching* & 0.0762 & 0.0000 & \second{22.0318} & 0.7395 & RefSTAR* & \second{0.6649} & 0.2005 & 0.4160 & 22.7635 & \second{0.2896} \\
DATSR* & 0.0735 & 0.0000 & 22.0063 & 0.7382 & InstantRestore* & 0.6515 & 0.2697 & 0.2759 & \second{22.9604} & 0.2930 \\
SUPIR+ReFIR* & 0.1339 & 0.0000 & 19.9177 & 0.4377 & ReF-LDM* & 0.5913 & \second{0.2857} & 0.1970 & 22.7719 & 0.3220 \\
SeeSR+ReFIR* & 0.1730 & 0.0000 & 20.8800 & 0.3134 & FaceMe* & 0.5761 & 0.2699 & \second{0.1508} & 21.8266 & 0.3168 \\
Ada-RefSR* & \second{0.1735} & 0.0000 & 21.5025 & \second{0.2553} & RestorerID* & 0.5032 & 0.2170 & 0.2074 & 22.5989 & 0.3554 \\
\midrule
\cellcolor{gray!15}\textbf{Ours*} & \cellcolor{gray!15}\best{0.8560} & \cellcolor{gray!15}\best{0.7625} & \cellcolor{gray!15}\best{23.4198} & \cellcolor{gray!15}\best{0.1797} & \cellcolor{gray!15}\textbf{Ours*} & \cellcolor{gray!15}\best{0.6675} & \cellcolor{gray!15}\best{0.3891} & \cellcolor{gray!15}\best{0.1287} & \cellcolor{gray!15}\best{23.8156} & \cellcolor{gray!15}\best{0.2716} \\
\bottomrule
\end{tabular}}

\vspace{3pt}
\centering
\includegraphics[width=\textwidth]{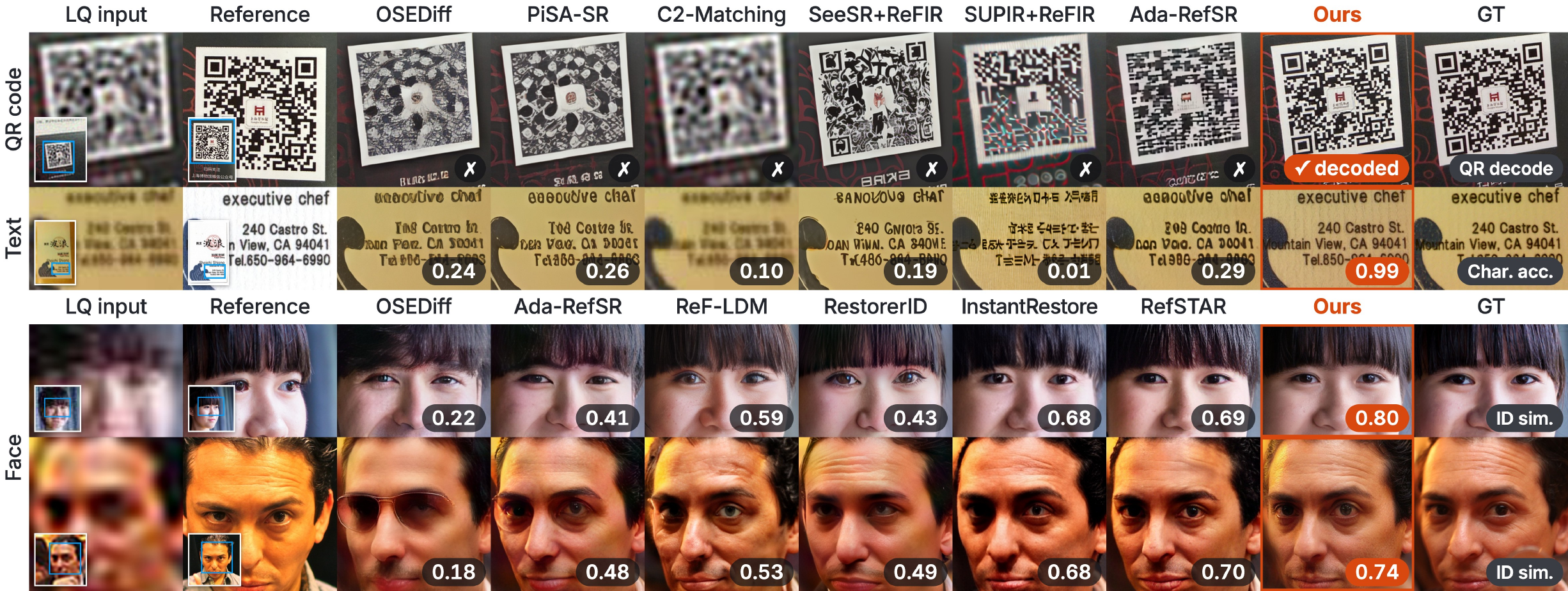}
    \captionof{figure}{Qualitative comparison on VeriRef with correct references. \emph{Top:} codes and text, with decoding success and character accuracy. \emph{Bottom:} faces, with identity similarity to the ground truth. Insets in the first two columns show the full LQ input and reference.}
    \label{fig:qual_content}
\end{figure*}

\subsection{Dual-Scale Reference Gating}
\label{sec:drg}

A restorer that reads its reference this directly also reproduces content that does not belong to the target, and the two pathways fail at different scales. Through the global pathway, an unrelated reference shifts the semantics of the whole image. Through the spatial pathway, target tokens retrieve similar-looking but mismatched content, such as digits of a different code or facial details of a different person, even when the reference is semantically relevant. DRG places one gate on each pathway, with no learned parameters and no training on wrong references (\cref{sec:training}).

\vspace{-4mm}
\paragraph{Semantic Relevance Gating (SRG).} SRG admits the reference Redux tokens only for a globally relevant reference:
\begin{equation}
g=\mathbbm{1}\!\left[\cos\!\big(f(\mI_L)-\vmu,\;f(\mI_R)-\vmu\big)\ge\tau\right],
\label{eq:gate}
\end{equation}
where $f$ is the mean SigLIP token feature and $\vmu$ its mean over training images; centering removes a direction shared by all images that makes unrelated pairs look similar. $\tau$ is the 10th percentile of scores on correct training pairs. Since SRG closes the global pathway only, local content can still be read through the spatial pathway.

\vspace{-4mm}
\paragraph{Bidirectional Consistency Gating (BCG).} A correct correspondence is consistent in both directions~\cite{cyclecorr}: if target token $i$ reads reference token $j$, then $j$ should attend back near $i$. Let
\begin{equation}
\bar{B}_{ji}=\frac1H\sum_{h=1}^{H}\softmax_{i'\in\mathcal T}\!\Big(\frac{(\vq^h_j)^\top \vk^h_{i'}}{\sqrt d}\Big)_{i}
\end{equation}
be the head-averaged reverse attention of reference token $j$ over the target tokens $\mathcal T$. BCG adds to the target-to-reference attention logits the bias
\begin{equation}
b_{ij}=\log\frac{\max_{i'\in\Omega(i)}\bar{B}_{ji'}}{\max_{i'\in\mathcal T}\bar{B}_{ji'}}\;\le 0,
\label{eq:mcb}
\end{equation}
where $\Omega(i)$ is the $7\times7$ neighborhood of $i$. A reference token whose strongest reverse match lies elsewhere, as with repeated patterns or mismatched content, is attenuated, shifting attention from the inconsistent match toward the target's own tokens and consistent reference content. BCG is applied in 15 of the 57 blocks, selected by layer-wise ablation on held-out training pairs.

\subsection{Training}
\label{sec:training}
We first adapt FLUX to single-image super-resolution with LoRA~\cite{lora} on the transformer and the LQ encoder, using online Real-ESRGAN degradations~\cite{realesrgan} and the flow-matching objective~\cite{flowmatching}. The single-image adapter is then merged, and a reference-conditioned adapter is trained on multi-view scene pairs, same-identity face pairs, and synthetic same-image pairs, with 10\% of samples left without a reference so that single-image behavior is retained. All training pairs share content with their targets; no mismatched references are used. Finally, we distill the model into a three-step sampler with latent adversarial distillation~\cite{ladd} combined with latent and perceptual reconstruction losses, simulating earlier sampler steps with the student itself so that training inputs match inference. Inference uses the noise levels $(1,0.75,0.25)$, predicting the clean latent at each step and re-noising it to the next level. A one-step variant is further distilled from the three-step model by regressing its deterministic output from the noised LQ input at $\sigma=0.9$ (latent MSE and LPIPS~\cite{lpips}, plus a ground-truth term), with the gates applied unchanged.

\section{VeriRef Benchmark}
\label{sec:benchmark}

Standard RefSR benchmarks measure full-image similarity, which a strong prior can improve without reading the reference. VeriRef (\cref{fig:benchmark}) follows three principles: (i) the evaluated content is legible in the ground truth but not in the LQ input, and selection never uses restoration outputs; (ii) each target is paired with a correct reference showing the same content and a wrong one showing different content of the same kind, so recovery and erroneous transfer are measured together; (iii) metrics test the content, not its appearance.

\noindent\textbf{Text.} 100 photographs of book covers, business cards, and CD and DVD covers from SMVS~\cite{smvs}, with 467 annotated regions scored by OCR character accuracy; regions with LQ character height below five pixels are marked hard.

\noindent\textbf{Codes.} 55 1D barcodes and 25 QR codes, recovered only if decoded to the exact payload. Under a wrong reference, we check whether the distractor's digits are copied.

\noindent\textbf{Faces.} 100 identities, each with a target degraded until its ArcFace~\cite{arcface} similarity to the ground truth falls below 0.45 while landmarks remain localizable, another photograph of the same person as the correct reference, and a different person as the wrong one. We report similarity to the ground truth and, under a wrong reference, to the distractor.

All targets use fixed second-order degradations, and all methods receive identical inputs and references.

\begin{figure*}[t]
\centering
\captionof{table}{Results on CUFED5 and WR-SR. * denotes reference-based methods. C2-Matching and DATSR are trained with bicubic degradations. \best{Red} and \second{blue}: best and second best on each dataset.}
\label{tab:general}
\vspace{-1.5mm}
\footnotesize
\setlength{\tabcolsep}{7.5pt}
\adjustbox{width=\textwidth}{%
\begin{tabular}{lll|ccccc|ccc}
\toprule
Dataset & Method & Venue & PSNR$\uparrow$ & SSIM$\uparrow$ & LPIPS$\downarrow$ & DISTS$\downarrow$ & FID$\downarrow$ & NIQE$\downarrow$ & MUSIQ$\uparrow$ & CLIPIQA$\uparrow$ \\
\midrule
\multirow{9}{*}{CUFED5} & C2-Matching* & CVPR 2021 & \second{20.7645} & 0.5179 & 0.7636 & 0.3600 & 283.15 & 9.2092 & 20.0918 & 0.1986 \\
 & DATSR* & ECCV 2022 & 20.7627 & 0.5178 & 0.7711 & 0.3550 & 283.64 & 9.3396 & 19.6824 & 0.1999 \\
\cmidrule{2-11}
 & Real-ESRGAN & ICCVW 2021 & 20.3078 & \second{0.5543} & 0.3698 & 0.2242 & 175.98 & 3.8883 & 64.7213 & 0.4207 \\
 & OSEDiff & NeurIPS 2024 & 20.3734 & 0.5355 & 0.3735 & 0.2136 & 156.24 & 4.4695 & 64.7508 & \best{0.5837} \\
 & PiSA-SR & CVPR 2025 & 20.3371 & 0.5286 & 0.3707 & 0.2102 & 164.63 & 4.1989 & 64.2835 & 0.5040 \\
 & SUPIR+ReFIR* & NeurIPS 2024 & 18.9256 & 0.4703 & 0.4292 & 0.2068 & 151.62 & 4.1561 & 63.0966 & 0.4423 \\
 & SeeSR+ReFIR* & NeurIPS 2024 & 20.2138 & 0.5263 & 0.3472 & 0.1868 & 137.36 & \second{3.7883} & \best{70.8899} & 0.5470 \\
 & Ada-RefSR* & ICLR 2026 & 20.4845 & 0.5462 & \second{0.2894} & \second{0.1632} & \second{128.03} & 3.8704 & 67.9183 & 0.5171 \\
\cmidrule{2-11}
 & \cellcolor{gray!15}\textbf{Ours*} & \cellcolor{gray!15}-- & \cellcolor{gray!15}\best{21.6406} & \cellcolor{gray!15}\best{0.6135} & \cellcolor{gray!15}\best{0.1945} & \cellcolor{gray!15}\best{0.1170} & \cellcolor{gray!15}\best{80.35} & \cellcolor{gray!15}\best{3.4124} & \cellcolor{gray!15}\second{69.0849} & \cellcolor{gray!15}\second{0.5490} \\
\midrule
\multirow{9}{*}{WR-SR} & C2-Matching* & CVPR 2021 & \best{22.5929} & 0.5679 & 0.7581 & 0.3272 & 159.44 & 9.4240 & 22.1701 & 0.2738 \\
 & DATSR* & ECCV 2022 & \second{22.5925} & 0.5674 & 0.7638 & 0.3247 & 157.73 & 9.5944 & 21.7863 & 0.2781 \\
\cmidrule{2-11}
 & Real-ESRGAN & ICCVW 2021 & 22.1487 & \best{0.5974} & 0.3631 & 0.2179 & 97.83 & \second{3.6977} & 64.5264 & 0.6076 \\
 & OSEDiff & NeurIPS 2024 & 21.6666 & 0.5640 & 0.3383 & 0.1893 & 67.43 & 4.1361 & 71.0550 & \second{0.7029} \\
 & PiSA-SR & CVPR 2025 & 21.8810 & 0.5682 & 0.3384 & 0.1957 & 79.38 & 4.0375 & 69.5223 & 0.6410 \\
 & SUPIR+ReFIR* & NeurIPS 2024 & 20.8785 & 0.5368 & 0.4086 & 0.2033 & 76.34 & 3.8291 & 67.2172 & 0.6253 \\
 & SeeSR+ReFIR* & NeurIPS 2024 & 21.8823 & 0.5689 & 0.3402 & 0.1862 & 61.77 & 3.9045 & \best{72.1101} & \best{0.7073} \\
 & Ada-RefSR* & ICLR 2026 & 21.9718 & 0.5776 & \second{0.3061} & \second{0.1610} & \second{53.37} & 3.7430 & 69.8632 & 0.6610 \\
\cmidrule{2-11}
 & \cellcolor{gray!15}\textbf{Ours*} & \cellcolor{gray!15}-- & \cellcolor{gray!15}22.1999 & \cellcolor{gray!15}\second{0.5886} & \cellcolor{gray!15}\best{0.2662} & \cellcolor{gray!15}\best{0.1359} & \cellcolor{gray!15}\best{37.54} & \cellcolor{gray!15}\best{3.5535} & \cellcolor{gray!15}\second{71.7558} & \cellcolor{gray!15}0.6641 \\
\bottomrule
\end{tabular}}

\vspace{3pt}
\centering
\includegraphics[width=\textwidth]{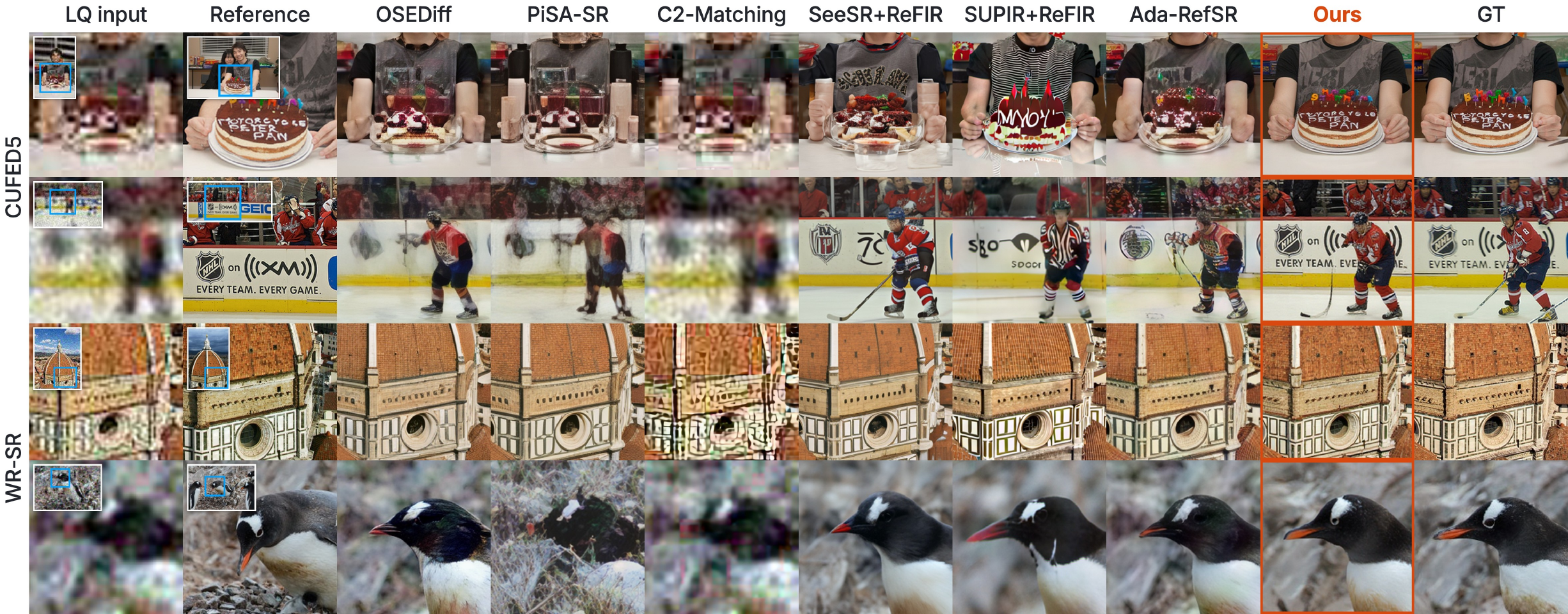}
    \vspace{-5mm}
    \captionof{figure}{Qualitative comparison on CUFED5 and WR-SR. Insets in the first two columns show the full LQ input and reference.}
    \label{fig:qual_scene}
\end{figure*}

\section{Experiments}
\label{sec:exp}

\subsection{Setup}
\label{sec:setup}
\vspace{-1mm}
\textbf{Implementation.} We use FLUX.1-Fill~\cite{fluxfill} with transformer and LQ-encoder LoRA ranks of 64 and 16. The single-image stage is trained on 49,107 LSDIR~\cite{lsdir} images and 10,000 FFHQ~\cite{ffhq} images. The reference stage samples MegaScenes~\cite{megascenes} multi-view pairs, CelebRef-HQ identity pairs~\cite{celebhq}, and synthetic LSDIR pairs with proportions of approximately 39\%, 14\%, and 37\%, plus 10\% single-image samples. All experiments perform $\times4$ super-resolution at native resolution ($\times8$ for VeriRef faces). Unless noted, all results use the three-step sampler with guidance 30 and DRG with both SRG ($\tau=0.172$) and BCG.

\vspace{-4mm}
\paragraph{Baselines.} We compare with single-image restorers (Real-ESRGAN~\cite{realesrgan}, OSEDiff~\cite{osediff}, PiSA-SR~\cite{pisasr}), general-scene RefSR methods (C2-Matching~\cite{jiang2021c2matching}, DATSR~\cite{cao2022datsr}, SeeSR~\cite{seesr} and SUPIR~\cite{supir} with ReFIR~\cite{guo2024refir}, Ada-RefSR~\cite{wang2026adarefsr}), and, on faces, reference-based face restorers (ReF-LDM~\cite{refldm}, RestorerID~\cite{restorerid}, InstantRestore~\cite{instantrestore}, FaceMe~\cite{faceme}, RefSTAR~\cite{refstar}). All baselines use their official releases with identical inputs and references (none for single-image methods); on CUFED5~\cite{zhang2019srntt} and WR-SR~\cite{jiang2021c2matching}, we follow the test protocol of Ada-RefSR~\cite{wang2026adarefsr} and evaluate all methods with the same code.

\subsection{How Precisely Is the Reference Read?}
\label{sec:exp_content}

On text and codes, the content that only the reference can supply, general-scene RefSR methods gain little from it: the best of them reaches a character accuracy of 0.174, against 0.131 for the single-image OSEDiff (\cref{tab:efficiency}) and none of them decodes a single code.

Our model recovers this content precisely. It reaches a character accuracy of 0.856 and decodes 61 of 80 codes, including 51 of 55 barcodes and 10 of 25 QR codes. On faces, it raises identity similarity from 0.474 for Ada-RefSR to 0.668, on par with the face-specific RefSTAR (0.665) and above InstantRestore (0.651). Under wrong references, however, these face restorers adopt the distractor: their similarity to the distractor rises to 0.416 and 0.276, and to the ground truth falls to 0.200 and 0.270. Our model keeps the distractor similarity at 0.129 and retains the highest ground-truth similarity, 0.389. It is the only method that both reads fine reference content and largely resists a wrong reference. Our outputs also obtain the lowest LPIPS on both text/codes and faces (\cref{tab:content}). \cref{fig:qual_content} shows representative examples.

\begin{figure*}[t]
\centering
\captionof{table}{How the reference enters. All variants share backbone, data, and our 1,500-step schedule, and are sampled with 30 steps without Redux, gating, or distillation. \emph{Right:} ID similarity during training. $^\ddagger$Ada-RefSR pathway on our backbone. $^\dagger$Extra extractor pass.}
\label{tab:ablation_inject}
\vspace{-1.5mm}
\renewcommand{\arraystretch}{1.05}
\newcommand{\injtab}{%
\begin{tabular}{ll|ccc|cc}
\toprule
& & \multicolumn{3}{c|}{VeriRef} & \multicolumn{2}{c}{LPIPS$\downarrow$} \\
Features & Attention & Text acc.$\uparrow$ & Decode$\uparrow$ & ID$\uparrow$ & CUFED5 & WR-SR \\
\midrule
\multicolumn{2}{l|}{No reference (single-image stage)} & 0.2588 & 0.0000 & 0.3589 & 0.3350 & 0.3124 \\
\midrule
Frozen extractor$^\dagger$ & Separate + gate$^\ddagger$ & 0.5273 & 0.0625 & 0.4730 & 0.2575 & 0.3038 \\
Backbone & Separate + fixed mix & 0.6998 & 0.6000 & 0.5266 & 0.2274 & 0.2983 \\
Backbone & Separate + gate & 0.7449 & 0.6125 & 0.5346 & 0.2382 & 0.2949 \\
Backbone & One-way & 0.7475 & 0.6500 & 0.5360 & 0.2267 & 0.2903 \\
\midrule
\rowcolor{gray!15}Backbone & Joint (ours) & \textbf{0.7491} & \textbf{0.7125} & \textbf{0.5625} & \textbf{0.2246} & \textbf{0.2901} \\
\bottomrule
\end{tabular}}
\newlength{\injR}\newlength{\injN}
\setlength{\tabcolsep}{4pt}%
\sbox{4}{\injtab}%
\sbox{0}{\adjustbox{width=0.60\textwidth}{\usebox{4}}}%
\sbox{2}{\includegraphics[height=\dimexpr\ht0+\dp0\relax]{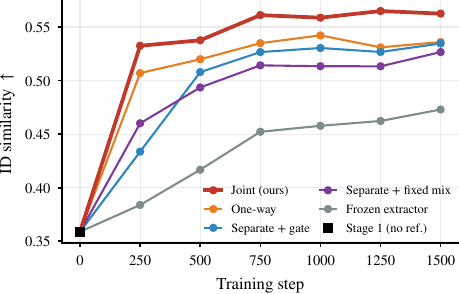}}%
\setlength{\injR}{\dimexpr\textwidth-\wd2-3mm\relax}%
\setlength{\injN}{\dimexpr\injR*\numexpr\wd4\relax/\numexpr\dimexpr0.60\textwidth\relax\relax\relax}%
\setlength{\tabcolsep}{\dimexpr4pt+(\injN-\wd4)/14\relax}%
\sbox{0}{\adjustbox{width=\injR}{\injtab}}%
\adjustbox{valign=c}{\usebox{0}}\hfill\adjustbox{valign=c}{\usebox{2}}
\end{figure*}

\begin{figure*}[t]
\begin{minipage}[t]{0.485\textwidth}
\vspace{0pt}\centering
\captionof{table}{Efficiency. Latency per $512\times512$ image on an RTX PRO 6000; VeriRef, correct references. \best{Red}/\second{blue}: best/second.}
\label{tab:efficiency}
\vspace{-1.5mm}
\setlength{\tabcolsep}{3pt}
\adjustbox{max width=\linewidth}{%
\begin{tabular}{l|c|cc|ccc}
\toprule
& & \multicolumn{2}{c|}{LPIPS$\downarrow$} & \multicolumn{3}{c}{VeriRef} \\
Method & Latency (s)$\downarrow$ & CUFED5 & WR-SR & Text acc.$\uparrow$ & Decode$\uparrow$ & ID$\uparrow$ \\
\midrule
OSEDiff & \best{0.101} & 0.3735 & 0.3383 & 0.1307 & 0.0000 & 0.3170 \\
Ada-RefSR* & \second{0.237} & 0.2894 & 0.3061 & 0.1735 & 0.0000 & 0.4743 \\
SeeSR+ReFIR* & 3.941 & 0.3472 & 0.3402 & 0.1730 & 0.0000 & 0.3619 \\
SUPIR+ReFIR* & 14.848 & 0.4292 & 0.4086 & 0.1339 & 0.0000 & 0.3330 \\
\midrule
Ours*, one step & 0.297 & \second{0.1972} & \best{0.2618} & \second{0.7523} & \second{0.7250} & \best{0.6700} \\
\rowcolor{gray!15}Ours*, three steps & 1.224 & \best{0.1945} & \second{0.2662} & \best{0.8560} & \best{0.7625} & \second{0.6675} \\
\bottomrule
\end{tabular}}
\end{minipage}\hfill
\begin{minipage}[t]{0.485\textwidth}
\vspace{0pt}\centering
\captionof{table}{Reference pathways. Spatial: reference image tokens in ATA. Global: reference Redux tokens. Best in bold.}
\label{tab:ablation_path}
\vspace{-1.5mm}
\setlength{\tabcolsep}{5pt}
\adjustbox{max width=\linewidth}{%
\begin{tabular}{cc|ccc|cc}
\toprule
& & \multicolumn{3}{c|}{VeriRef} & \multicolumn{2}{c}{LPIPS$\downarrow$} \\
Spatial & Global & Text acc.$\uparrow$ & Decode$\uparrow$ & ID$\uparrow$ & CUFED5 & WR-SR \\
\midrule
 & & 0.2850 & 0.0000 & 0.4169 & 0.3102 & 0.2960 \\
 & \checkmark & 0.4581 & 0.0000 & 0.4272 & 0.3012 & 0.2961 \\
\checkmark & & 0.7909 & 0.7500 & \textbf{0.6695} & 0.1954 & 0.2665 \\
\midrule
\rowcolor{gray!15}\checkmark & \checkmark & \textbf{0.8560} & \textbf{0.7625} & 0.6675 & \textbf{0.1945} & \textbf{0.2662} \\
\bottomrule
\end{tabular}}
\end{minipage}
\end{figure*}

\subsection{Comparison on RefSR Benchmarks}
\label{sec:exp_general}

\textbf{Benchmark results.} On the established benchmarks (\cref{tab:general}), our model obtains the best LPIPS~\cite{lpips}, DISTS~\cite{dists}, FID~\cite{fid}, and NIQE~\cite{niqe} on both datasets, reducing FID from 128.0 to 80.4 on CUFED5 and from 53.4 to 37.5 on WR-SR relative to Ada-RefSR, as well as the highest PSNR and SSIM~\cite{ssim} on CUFED5 and the second-highest MUSIQ~\cite{musiq} on both datasets (\cref{fig:qual_scene}). These results show that reading the reference closely does not cost general restoration quality.

\begin{figure}[t]
    \centering
    \includegraphics[width=0.6\linewidth]{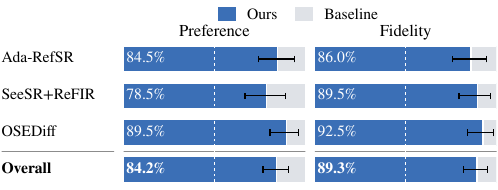}
    \caption{User study on CUFED5 and WR-SR: share of judgments choosing our result over each baseline, for preference and fidelity.}
\label{fig:user_study}
\end{figure}

\vspace{-4mm}
\paragraph{User study.} 
We use 20 images each from CUFED5 and WR-SR (\cref{fig:user_study}) and compare our result with Ada-RefSR, SeeSR+ReFIR, and OSEDiff in two tasks. For preference, participants see the LQ input and the reference and choose the restoration they prefer; for fidelity, they see the ground truth and choose the result whose content and details agree with it more closely. Twenty participants give 1,200 judgments, five per question. Our results are chosen in 84.2\% of preference and 89.3\% of fidelity judgments, and every 95\% interval lies above 50\%.

\vspace{-4mm}
\paragraph{Efficiency.} \cref{tab:efficiency} compares latency and quality. Our one-step variant already outperforms every baseline on all metrics in \cref{tab:efficiency}, at 0.30\,s per image, faster than the multi-step SeeSR+ReFIR and SUPIR+ReFIR. Its main weakness is fine reference content: text accuracy drops from 0.856 to 0.752, likely because details are largely written in the later, low-noise steps. Balancing quality and cost, our main model is the three-step one, which adds 0.93\,s per image.

\subsection{How Should the Reference Enter?}
\label{sec:inject}

The gap in \cref{tab:content} raises the question of how the reference should enter a restorer. \cref{tab:ablation_inject} compares reference pathways that share backbone, data, and schedule and differ only in where the reference features come from and how they are combined with the target; the second row reproduces the Ada-RefSR pathway on our FLUX backbone, and the following rows move it step by step toward joint attention. Three findings follow. First, a frozen extractor severely impairs reading fine content: the Ada-RefSR pathway decodes 6.3\% of the codes against 71.3\% with joint attention, so a stronger backbone alone does not explain the gap. Second, even with backbone features, how they are combined matters: separating the target and reference softmax lowers decoding to 60--61\%, and one-way injection to 65\%, each with lower identity similarity. Third, joint attention reads the reference best and learns it fastest: it is best on all metrics, needs no extra extractor pass, and leads in identity similarity at every checkpoint (0.53 vs.\ 0.43 for separate softmax after 250 steps), consistent with the pretrained attention already performing the matching.

\begin{table*}[t]
\centering
\caption{DRG ablation. \emph{Left:} wrong references; Digit copy: rate of 1D codes whose digits are copied from the wrong reference; Excess: LPIPS increase over reference-free outputs. \emph{Right:} correct references. Best in bold.}
\label{tab:ablation_gate}
\vspace{-2mm}
\setlength{\tabcolsep}{5pt}
\adjustbox{max width=\textwidth}{%
\begin{tabular}{cc|ccc|cc|cc|cc}
\toprule
& & \multicolumn{5}{c|}{Wrong reference} & \multicolumn{4}{c}{Correct reference} \\
& & \multicolumn{3}{c|}{VeriRef} & \multicolumn{2}{c|}{Excess$\downarrow$} & \multicolumn{2}{c|}{VeriRef} & \multicolumn{2}{c}{LPIPS$\downarrow$} \\
SRG & BCG & Digit copy$\downarrow$ & ID$_{\text{wrong}}\uparrow$ & ID leak$_{\text{wrong}}\downarrow$ & CUFED5 & WR-SR & Text acc.$\uparrow$ & Decode$\uparrow$ & CUFED5 & WR-SR \\
\midrule
 & & 0.2182 & 0.3700 & 0.1650 & 0.0109 & 0.0156 & 0.8584 & \textbf{0.7875} & \textbf{0.1922} & 0.2722 \\
\checkmark & & 0.2000 & 0.3716 & 0.1676 & 0.0055 & 0.0089 & 0.8520 & \textbf{0.7875} & \textbf{0.1922} & 0.2719 \\
 & \checkmark & \textbf{0.1091} & 0.3875 & \textbf{0.1263} & 0.0085 & 0.0095 & \textbf{0.8605} & 0.7625 & 0.1949 & 0.2667 \\
\midrule
\rowcolor{gray!15}\checkmark & \checkmark & \textbf{0.1091} & \textbf{0.3891} & 0.1287 & \textbf{0.0029} & \textbf{0.0034} & 0.8560 & 0.7625 & 0.1945 & \textbf{0.2662} \\
\bottomrule
\end{tabular}}
\end{table*}

\begin{figure}[t]
    \centering
    \includegraphics[width=0.9\linewidth]{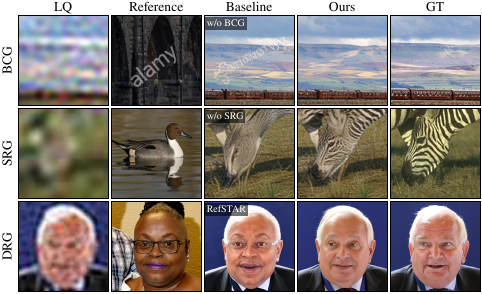}
    \vspace{-3mm}
    \caption{Effect of DRG. \emph{Top:} correct reference; BCG stops repeated structures from being matched to wrong locations. \emph{Middle:} wrong reference; SRG keeps its semantics out of the output. \emph{Bottom:} wrong reference; RefSTAR adopts the distractor identity, while our full model keeps the target identity.}
    \label{fig:gate_cmp}
\end{figure}

\begin{figure}[t]
\centering
\captionof{table}{Reference shape. All layouts enlarge the reference to the target area; our default (\cref{sec:ata}) keeps its native scale, giving a shorter sequence. Best in bold.}
\vspace{-3mm}
\label{tab:ablation_shape}
\setlength{\tabcolsep}{4pt}
\adjustbox{max width=\linewidth}{%
\begin{tabular}{l|ccc|cc}
\toprule
& \multicolumn{3}{c|}{VeriRef} & \multicolumn{2}{c}{LPIPS$\downarrow$} \\
Layout & Text acc.$\uparrow$ & Decode$\uparrow$ & ID$\uparrow$ & CUFED5 & WR-SR \\
\midrule
Diptych, crop & 0.8592 & 0.7250 & 0.6660 & 0.1966 & 0.2723 \\
Diptych, resize & 0.8968 & 0.7375 & 0.6660 & 0.1958 & 0.2710 \\
\midrule
\rowcolor{gray!15}Free shape (ATA) & \textbf{0.8992} & \textbf{0.7625} & \textbf{0.6675} & \textbf{0.1943} & \textbf{0.2699} \\
\bottomrule
\end{tabular}}
\end{figure}

\subsection{Ablation Studies}
\label{sec:ablation}

\paragraph{Asymmetric shapes.} We compare three layouts at the same reference token budget (\cref{tab:ablation_shape}): cropping or resizing the reference into the second half of a diptych, and keeping its own shape as in ATA. Cropping discards reference content and lowers text accuracy from 0.899 to 0.859; resizing distorts its geometry and decodes fewer codes. Keeping the shape is best on all metrics. Enlarging small references (0.59 of the target area on average for text) raises text accuracy further (0.899 vs.\ 0.856) but lengthens the sequence; we keep the native scale.

\vspace{-4mm}
\paragraph{Spatial and global pathways.} The spatial pathway carries the content (\cref{tab:ablation_path}): it is the only one that decodes codes or raises identity similarity substantially. The global pathway alone improves text partly and recovers no code, but added to the spatial pathway it raises text accuracy from 0.791 to 0.856, consistent with its role of locating content.

\vspace{-4mm}
\paragraph{Dual-Scale Reference Gating.} The two gates of DRG act at different scales (\cref{tab:ablation_gate}). BCG addresses local content: it reduces the rate at which distractor digits are copied from 0.218 to 0.109 and the similarity to a distractor identity from 0.165 to 0.126, while SRG alone barely changes these. SRG addresses global transfer on general scenes, where a wrong reference shifts the whole image: on top of BCG, it removes 66\% and 64\% of the excess LPIPS on CUFED5 and WR-SR. With both SRG and BCG, wrong references raise LPIPS by at most 0.0034 over using no reference. With correct references, the gates leave text accuracy and LPIPS nearly unchanged, while BCG costs two decoded codes. Examples are shown in \cref{fig:gate_cmp}.

\section{Conclusion}
\label{sec:conclusion}

We presented PreciseRef, which reads fine reference content and resists misleading references. Asymmetric Twin-Attention lets target and reference tokens share joint attention, and Dual-Scale Reference Gating filters irrelevant context and inconsistent matches without extra parameters or mismatched-reference training. On VeriRef, PreciseRef raises character accuracy from 17.4\% to 85.6\% and decodes 76.3\% of codes, which no prior method decodes, with state-of-the-art perceptual quality on CUFED5 and WR-SR. These results suggest that RefSR should be judged by what it recovers from the reference, not by plausibility alone.
\clearpage
\section*{LLM Usage Statement}

We used ChatGPT (OpenAI) and Claude (Anthropic) for language polishing, \LaTeX{} formatting, reviewer-style feedback on the manuscript, and assistance with implementation code, which the authors reviewed and verified. All scientific contributions, methodological decisions, experimental results, and conclusions were developed and verified by the authors.

{\small
\bibliographystyle{ieeenat_fullname}
\bibliography{main}}

\end{document}